\documentclass[11pt]{article}

\usepackage[preprint]{style}
\usepackage[utf8]{inputenc} 
\usepackage[T1]{fontenc}    
\usepackage{hyperref}       
\usepackage{url}            
\usepackage{booktabs}       
\usepackage{amsfonts}       
\usepackage{nicefrac}       
\usepackage{microtype}      
\usepackage{xcolor}         
\usepackage{amsmath}
\usepackage{graphicx}       
\usepackage{tikz}           
\usetikzlibrary{positioning}
\usepackage{caption}

\title{FlowMPC: Improving Flow Matching policies with World Models}

\author{%
  Chandon Hamel \\
  Stanford University\\
  \texttt{chandonh@stanford.edu} \\
}

\begin{document}

\maketitle

\begin{abstract}
Flow Matching (FM) is a powerful approach for behavior cloning in multimodal action spaces \citep{pmlr-v305-jiang25a}, but because it is not trained to directly maximize expected return, there is still room to improve how FM policies act at test time. This work investigates whether a learned world model can improve FM policies by enabling Model Predictive Path Integral (MPPI) planning over candidate action sequences proposed by the policy. Building on TD-MPC2 \citep{hansen2024tdmpc2scalablerobustworld}, I introduce \texttt{FlowMPC}, a framework that combines an imitation-learned FM policy with a learned world model for test-time planning in ManiSkill manipulation tasks \citep{taomaniskill3}. Across \texttt{PickCube} and \texttt{PickSingleYCB}, adding the world model improved performance over the FM policy alone, with especially clear gains in end-of-episode success. These results suggest that world-model-based planning can effectively complement flow-based imitation policies without modifying the FM training objective.
\end{abstract}

\section{Introduction}

Flow Matching (FM) has recently emerged as a powerful approach for behavior cloning in multimodal action spaces \citep{pmlr-v305-jiang25a}. That makes it especially appealing for robotic manipulation, where policies often have to act from high-dimensional visual observations and where there may be many different action trajectories that successfully solve the same task. In settings like these, expressive generative policy classes can be a much better fit than simple regression-style behavior cloning.

At the same time, FM is still an imitation learning method. It is trained to match expert behavior, not to directly maximize expected return, so its performance is ultimately limited by the quality and coverage of the demonstration data. In practice, that can lead to policies that look strong on in-distribution states but struggle to recover from small errors, compounding mistakes, or states that were not well represented in the expert dataset. Recent work has started to address this by improving flow-based policies with reinforcement learning \citep{mcallister2026flow,pfrommer2025reinforcementlearningflowmatchingpolicies}, but those approaches do so by changing the learning objective itself.

This work asks a different question: can a learned world model improve an FM policy without changing the FM objective? Rather than refining the policy through reinforcement learning, I use the FM policy as a proposal mechanism for test-time planning. The idea is to let the FM policy do what it is already good at---generate rich, multimodal action trajectories---and then use a world model to evaluate which of those trajectories are actually promising.

To do this, I build on TD-MPC2 \citep{hansen2024tdmpc2scalablerobustworld} and replace its learned actor with an imitation-learned FM policy inside an MPPI-based planning loop. That sounds simple at a high level, but making the two work together required several changes to the original TD-MPC2 setup. In particular, I added a behavior-cloning action predictor for terminal value bootstrapping and, for one of the tasks, used a separate state-space version of the world model that was better suited to the setting. Additionally, the reward and value functions were adapted to complement the framework of the learning objectives. These changes made it possible to plan over candidate trajectories proposed by the FM policy while keeping action selection closer to the policy's learned distribution and accurately predict future outcomes.

I evaluated this approach in ManiSkill on two robotic manipulation tasks, \texttt{PickCube-v1} and \texttt{PickSingleYCB-v1}, using expert trajectories collected from a SAC policy with privileged state information \citep{taomaniskill3,haarnoja2018softactorcriticoffpolicymaximum}. Across both tasks, augmenting the FM policy with a learned world model and MPPI planning improves performance over the FM policy alone, with the clearest gains appearing in end-of-episode success. End success improves from 93.14\% to 97.44\% on \texttt{PickCube} and from 56.81\% to 66.41\% on \texttt{PickSingleYCB}, suggesting that planning is especially useful for correcting small errors and maintaining stable control through the end of the episode.

More broadly, this work explores a simple hybrid between imitation learning and model-based reinforcement learning: use imitation to learn a strong multimodal action prior, then use planning to improve decisions at inference time. The scope here is limited to two tasks, but the results suggest that world-model-based planning can complement flow-based imitation policies in a meaningful way without requiring a new FM training objective.

\section{Related Work}

\paragraph{Flow-based policies for imitation learning.}
Recent work has shown that flow-based generative modeling can be an effective approach for policy learning in continuous, multimodal action spaces. In particular, FM provides a flexible framework for learning action distributions and has proven especially appealing for robotic behavior cloning, where multiple distinct action sequences may successfully solve the same task \citep{pmlr-v305-jiang25a}. Compared to more conventional unimodal regression-based behavior cloning, these methods are better suited to capturing the diversity and trajectory-level structure of expert demonstrations.

At the same time, flow-based policies inherit a key limitation of imitation learning: they are trained to reproduce expert behavior rather than to directly optimize task return. As a result, their performance is bounded by the quality and coverage of the demonstration data, and they may struggle to recover once the system drifts away from states represented in the expert dataset. This limitation is particularly relevant in robotic manipulation, where small action errors can compound over time and lead to failed task completion.

\paragraph{Improving flow policies with reinforcement learning.}
A natural way to overcome the limitations of pure imitation is to refine flow-based policies with reinforcement learning. Recent work has begun to explore this direction by introducing reinforcement learning objectives for FM policies, showing that flow-based action generators can be adapted beyond standard behavior cloning \citep{mcallister2026flow,pfrommer2025reinforcementlearningflowmatchingpolicies}. These methods are closely related to the motivation of this work, since they share the goal of improving FM policies beyond what is possible through expert imitation alone.

However, unlike those approaches, this work does not modify the FM training objective or optimize the policy directly with reinforcement learning. Instead, it treats the learned FM policy as a strong action prior and uses model-based planning at inference time to search over candidate trajectories proposed by that policy. In this sense, the work is complementary to prior RL-based FM refinement methods. Rather than improving the generator through new policy-learning objectives, it improves decision making through test-time planning.

\paragraph{World models and model-based planning.}
This work also builds on the literature on learned world models for continuous control, especially TD-MPC2 \citep{hansen2024tdmpc2scalablerobustworld}. TD-MPC2 combines a learned dynamics model, reward model, and value estimator with planning in latent space, yielding a scalable model-based reinforcement learning framework for high-dimensional continuous-action problems. In the standard TD-MPC2 setup, candidate actions are proposed by a learned actor and refined through planning.

My approach borrows the world-model-and-planning backbone of TD-MPC2, but replaces the actor-centered action proposal mechanism with trajectories sampled from an imitation-learned FM policy. This change is motivated by the observation that FM policies can provide high-quality, multimodal trajectory proposals, while the world model can evaluate and refine those proposals according to predicted task return. Adapting TD-MPC2 to this setting required several nontrivial modifications, including a behavior-cloning action predictor for end-of-horizon bootstrapping, changes to the reward and value functions, and, for one task, a state-space variant of the world model.

\paragraph{Positioning of this work.}
Taken together, prior work suggests two promising ingredients for robust policy improvement: expressive generative policies that capture multimodal expert behavior, and world models that enable lookahead planning. This work asks whether a FM policy can be improved not by changing how it is trained, but by coupling it with a learned world model that plans over the action sequences the policy already knows how to generate.

To my knowledge, this specific combination of FM-based imitation learning and TD-MPC2-style world-model planning has been relatively underexplored. The contribution of this work is therefore not a new standalone imitation learning algorithm or a new model-based RL objective, but rather a hybrid framework that connects these two lines of work and evaluates whether the combination can improve manipulation performance in practice.

\section{Method}

This section describes the overall training paradigm used in FlowMPC, the architectural modifications made to TD-MPC2 to support a FM policy, and the training objectives used to fit the world model. The key idea is to train a FM policy from expert demonstrations, then use that policy to propose candidate action sequences that are evaluated and refined with a learned world model and MPPI planning.

\subsection{Overall Training Paradigm}

The full training pipeline consists of five stages. First, I trained a SAC expert with access to privileged state information in order to obtain a reliable source of expert demonstrations \citep{haarnoja2018softactorcriticoffpolicymaximum}. Second, I collected full trajectories containing RGB observations, proprioceptive state, actions, and rewards. These first two stages can be any form of expert demonstration collection. Using an expert SAC agent suited this work's chosen environment. Third, I trained a FM policy on those expert trajectories \citep{pmlr-v305-jiang25a}. Fourth, I recollected the expert trajectories with RGB observations preprocessed by the FM policy's visual encoder, producing compact visual latents for downstream world model training (encoding previously collected data is also possible). Finally, I trained the world model online with the replay buffer preloaded from the expert data.

This preprocessing step is important because it allows the world model to reuse the FM policy's visual representation instead of learning directly from raw images. In practice, the visual encoder produces a compact visual latent \(z_{\mathrm{viz},t}\), which is then combined with the robot state before being passed into the world model:
\[
z_{\mathrm{viz},t} = \mathrm{VisualEncoder}(o^{\mathrm{RGB}}_t), 
\qquad
x_t = [s_t, z_{\mathrm{viz},t}].
\]
This makes the world model substantially easier to train, and it is especially useful for the state-space variant described below, where predicting future RGB observations directly would be impractical.

At inference time, the FM policy is used as a structured proposal distribution inside an MPPI planner. Given the current observation, the planner samples \(K\) candidate action trajectories, with \(K_\pi\) trajectories proposed by the FM policy and the remaining \(K-K_\pi\) trajectories sampled randomly. Each trajectory is rolled out in the learned world model, scored using predicted rewards and a terminal value estimate, and then refined using MPPI before the first action is executed. In the final configuration, \(K=512\) and \(K_\pi=24\) \citep{hansen2024tdmpc2scalablerobustworld}. For efficiency, the FM visual encoding is computed once per observation and reused across all FM samples and Euler integration steps at inference time.

\begin{figure}[t]
\centering
\begin{tikzpicture}[
node distance=1.2cm and 0.8cm,
every node/.style={font=\small},
box/.style={draw, rounded corners, align=center, minimum width=3.1cm, minimum height=1.1cm},
arrow/.style={->, thick}
]
\node[box] (obs) {Observation\\$o_t$};
\node[box, right=of obs] (policy) {Propose action sequences\\$K_\pi$ from $\pi_{\mathrm{FM}}$\\$K-K_\pi$ random};
\node[box, right=of policy] (wm) {World model rollout\\$\hat{x}, \hat{r}$};
\node[box, below=of wm] (score) {Score trajectories\\$\sum_h \hat{r}_{t+h} + \hat{Q}(\hat{x}_{t+H+1}, \pi_{\mathrm{BC}}(\hat{x}_{t+H+1}))$};
\node[box, left=of score] (mppi) {MPPI refinement\\select action $a_t$};
\node[box, left=of mppi] (env) {Execute action\\$a_t$ in environment};

\draw[arrow] (obs) -- (policy);
\draw[arrow] (policy) -- (wm);
\draw[arrow] (wm) -- (score);
\draw[arrow] (score) -- (mppi);
\draw[arrow] (mppi) -- (env);
\draw[arrow] (env) -- (obs);
\end{tikzpicture}
\caption{Planning with a FM policy and learned world model. The FM policy proposes a subset of candidate action trajectories, which are evaluated and refined using model-based rollouts and MPPI.}
\label{fig:flowmpc_pipeline}
\end{figure}
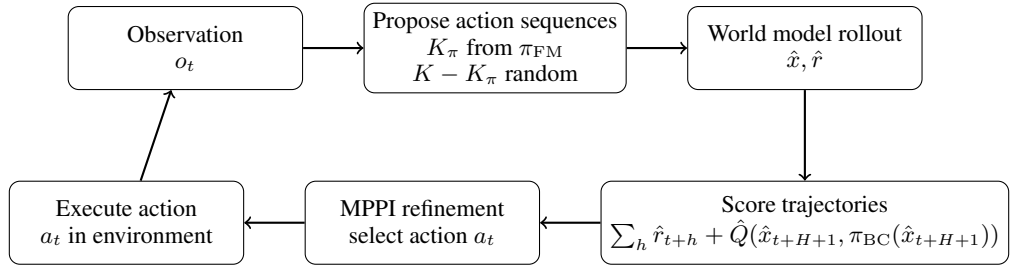

\subsection{Flow Matching Policy Architecture}

The Flow Matching policy used in this work is implemented as a conditional 1D temporal U-Net over action trajectories. Given the current observation and a flow timestep, the network predicts a vector field over an action sequence rather than a single action, which makes it well suited to multimodal trajectory generation. The model is conditioned on proprioceptive state and, when RGB observations are available, on a learned visual latent produced by a separate visual encoder \citep{pmlr-v162-janner22a}.

\subsection{TD-MPC2 Adaptations}

As a starting point, the world model is based on TD-MPC2, which learns a representation model, dynamics model, reward model, and action-conditioned critic model. In standard TD-MPC2, a learned actor proposes actions rolled out autoregressively with the dynamics model that are then optimized through planning. In FlowMPC, that actor is replaced by an imitation-learned FM policy that generates full action sequences, and the world model is used to evaluate candidate trajectories proposed by the FM policy rather than to train a policy through actor-critic updates \citep{hansen2024tdmpc2scalablerobustworld}.

This change required several modifications to the original architecture. First, I added a behavior cloning action predictor, denoted \(\pi_{\mathrm{BC}}\), that maps the current model state to an action:
\[
\hat{a}_t = \pi_{\mathrm{BC}}(x_t).
\]
This network is used both during training and at the end of planning rollouts, where it provides an in-distribution action for terminal value bootstrapping. Without this head, the value model would need to evaluate terminal states paired with arbitrary sampled actions, which made planning less stable.

Second, I modified the representation interface so that the world model consumes the FM visual latent directly together with the proprioceptive state, rather than maintaining separate visual and state encoders whose outputs are later concatenated. Concretely, the FM visual encoder first produces \(z_{\mathrm{viz},t}\), and the world model then operates on the combined input \(x_t = [s_t, z_{\mathrm{viz},t}]\). This simplification made it easier to reuse the FM policy's learned visual features, allowed the representation model to fully leverage the frozen encoder's output, and reduced redundancy between the imitation policy and the world model.

Third, I simplified the reward and value prediction targets relative to the original TD-MPC2 formulation. TD-MPC2 predicts discretized symlog-transformed rewards and values, which is a sensible design choice for the broad multitask settings it targets \citep{hansen2024tdmpc2scalablerobustworld}. In this work, that machinery was unnecessary. Because the environments considered here are single-task manipulation problems with binary rewards, I instead used a binary cross-entropy loss for reward prediction and a standard mean-squared-error objective for Q-value regression. This made the value-learning setup better matched to the scale and structure of the tasks and kept the training objective considerably simpler.

Fourth, I implemented a state-space variant of the world model in addition to the latent TD-MPC2-style version. The latent version follows the standard TD-MPC2 design more closely: it uses a learned representation, supports multi-step rollouts, and is trained from temporally contiguous slices sampled from the replay buffer. By contrast, the state-space version omits the learned latent encoder and instead predicts directly in the space of concatenated proprioceptive state and FM visual latent. This removes the shared upstream representation module, which in turn allows the dynamics, reward, value, and behavior-cloning networks to be optimized without all gradients flowing through a common encoder. It also used a simpler one-step training setup based on standard single-transition tuples \((x_t, a_t, r_t, x_{t+1})\), similar to SAC-style replay \citep{haarnoja2018softactorcriticoffpolicymaximum}.

These two variants served different purposes in the final system. For \texttt{PickCube}, the state-space world model with planning horizon \(H=1\) performed best. For \texttt{PickSingleYCB}, a latent-space model with \(H=3\) gave better results, suggesting that the more difficult task benefited from a more expressive learned representation and longer planning horizon.

\subsection{Training Objectives}

Let \(x_t\) denote the model state used by the world model. In the state-space version, \(x_t\) is the true state representation formed by concatenating proprioceptive state and FM visual latent, \(x_t = [s_t, z_{\mathrm{viz},t}]\). In the latent version, \(x_t\) instead denotes the learned latent state produced by the representation model. This notation allows the same update equations to be written for both versions of the world model.

The world model is trained from replay using separate objectives for dynamics prediction, reward prediction, value estimation, and the behavior-cloning action head. For a transition \((x_t, a_t, r_t, x_{t+1})\), the dynamics model predicts the next model state,
\[
\hat{x}_{t+1} = f_{\mathrm{dyn}}(x_t, a_t),
\]
and the reward model predicts the binary task reward,
\[
\hat{r}_t = f_{\mathrm{rew}}(x_t, a_t).
\]
The value model consists of a set of \(Q\)-functions trained toward a one-step TD target constructed from the target networks and the behavior-cloning policy:
\[
y_t = r_t + \gamma \, \hat{Q}_{\mathrm{targ}}\!\left(x_{t+1}, \pi_{\mathrm{BC}}(x_{t+1})\right),
\]
where \(\hat{Q}_{\mathrm{targ}}\) denotes the minimum over the target \(Q\)-ensemble.

The resulting losses are
\[
\mathcal{L}_{\mathrm{dyn}} = \left\| \hat{x}_{t+1} - x_{t+1} \right\|_2^2,
\]
\[
\mathcal{L}_{\mathrm{rew}} = \mathrm{BCEWithLogits}(\hat{r}_t, r_t),
\]
\[
\mathcal{L}_{Q} = \sum_{i=1}^{N_Q} \left\| Q_i(x_t, a_t) - y_t \right\|_2^2,
\]
\[
\mathcal{L}_{\mathrm{BC}} = \left\| \pi_{\mathrm{BC}}(x_t) - a_t \right\|_2^2.
\]
The binary cross-entropy reward loss is used because the ManiSkill tasks considered here were set to provide binary rewards \citep{taomaniskill3}.

In the state-space implementation, these losses are applied directly to one-step replay transitions. In the latent implementation, training instead uses temporally contiguous slices and unrolls the model across the planning horizon. A consistency loss matches predicted latent rollouts to encoder-produced target latents, while the reward, value, and behavior-cloning losses are applied at each step under a down-weighting schedule across the horizon. Since all of these objectives depend on the shared encoded state, they jointly shape the representation model in the latent variant, unlike in the state-space version where the heads are decoupled from any learned encoder \citep{hansen2024tdmpc2scalablerobustworld}.

\subsection{Planning Objective}

At test time, the planner evaluates each candidate action sequence by rolling it forward in the learned world model and summing predicted rewards plus a terminal value estimate:
\[
J^{(k)} = \sum_{h=0}^{H} \hat{r}^{(k)}_{t+h}
+ \hat{Q}\!\left(\hat{x}^{(k)}_{t+H+1}, \pi_{\mathrm{BC}}(\hat{x}^{(k)}_{t+H+1})\right).
\]
MPPI then reweights and refines the sampled trajectories according to these scores, and the first action of the optimized sequence is executed in the environment. This procedure allows the agent to preserve the multimodal action prior learned by the FM policy while still selecting trajectories that are predicted to produce higher return under the world model \citep{hansen2024tdmpc2scalablerobustworld}.

\section{Experimental Setup}

\subsection{Environments and Observations}

I evaluate FlowMPC on two robotic manipulation tasks from ManiSkill: \texttt{PickCube-v1} and \texttt{PickSingleYCB-v1}. In both tasks, a robotic arm must pick up a target object, move it to a goal location, and remain sufficiently still to satisfy the success condition. These tasks provide a useful testbed for the proposed method because they require both coarse motion planning and fine-grained control near task completion \citep{taomaniskill3}.

Observations consist of RGB images together with proprioceptive robot state, and actions are continuous robot control commands including arm motion and gripper actuation. Rewards are binary: the agent receives a reward of \(1\) when the task success condition is met and \(0\) otherwise \citep{taomaniskill3}.

\subsection{Compared Methods and Evaluation}

The primary comparison is between the imitation-learned FM policy alone and the same FM policy augmented with a learned world model and MPPI planning. The goal is to measure whether the world model improves decision making at inference time beyond what the FM policy can achieve by imitation alone.

Evaluation is performed over 50-step episodes using two metrics: \textit{anytime success rate} and \textit{end success rate}. Anytime success measures whether the task succeeds at any point during the episode, while end success measures whether the agent is still successful at the final timestep. Reporting both metrics is useful in these manipulation tasks because it distinguishes between policies that can briefly complete the task and policies that can complete it while maintaining stable control through the end of the episode \citep{taomaniskill3}.

\subsection{Implementation Choices}

The final system uses different world model variants for the two tasks. For \texttt{PickCube}, the best-performing configuration uses the state-space world model with planning horizon \(H = 1\). For \texttt{PickSingleYCB}, the best-performing configuration uses the latent-space world model with planning horizon \(H = 3\). In both cases, the opposite configuration performed worse in preliminary experiments.

The \texttt{PickCube} world model uses hidden dimension 512, while the \texttt{PickSingleYCB} world model uses hidden dimension 384; both use two hidden layers. A cosine annealing learning rate schedule is used in the final configurations, and this had a particularly large effect on final \texttt{PickCube} performance. During planning, \(24\) of \(512\) candidate MPPI trajectories are proposed directly by the FM policy, with the remainder sampled randomly.

\section{Results}

\begin{figure}[h]
      \begin{center}
        \textcolor{blue}{\textbf{------ FM Policy baseline ------}}
      \end{center}
      \centering

      \begin{minipage}[t]{0.48\linewidth}
        \centering
        \includegraphics[width=\linewidth]{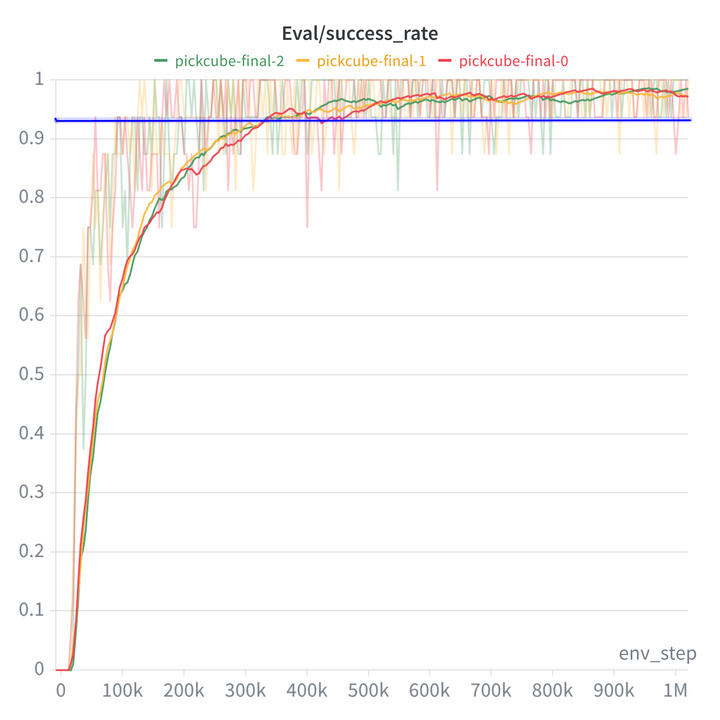}
        
        {\small \textbf{PickCube: End}}
      \end{minipage}
      \hfill
      \begin{minipage}[t]{0.48\linewidth}
        \centering
        \includegraphics[width=\linewidth]{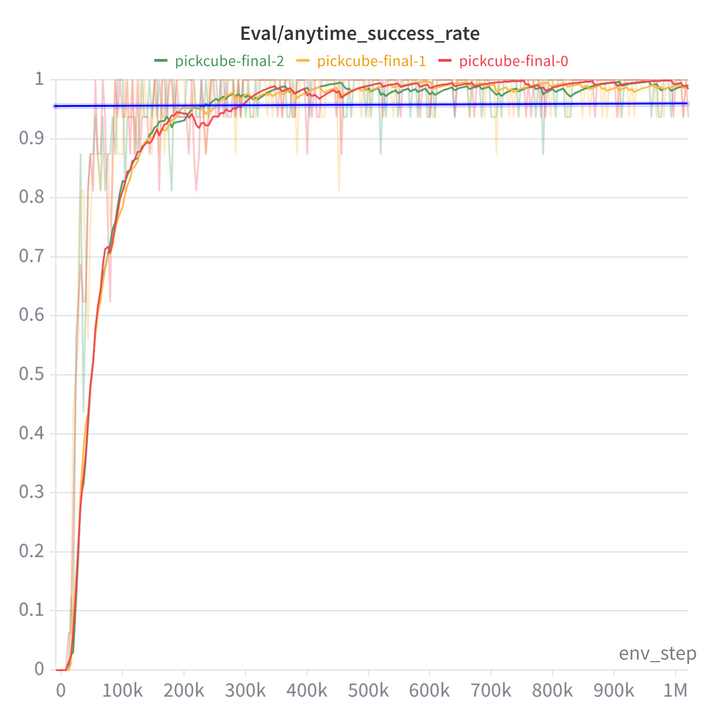}
        
        {\small \textbf{PickCube: Anytime}}
      \end{minipage}

      \begin{minipage}[t]{0.48\linewidth}
        \centering
        \includegraphics[width=\linewidth]{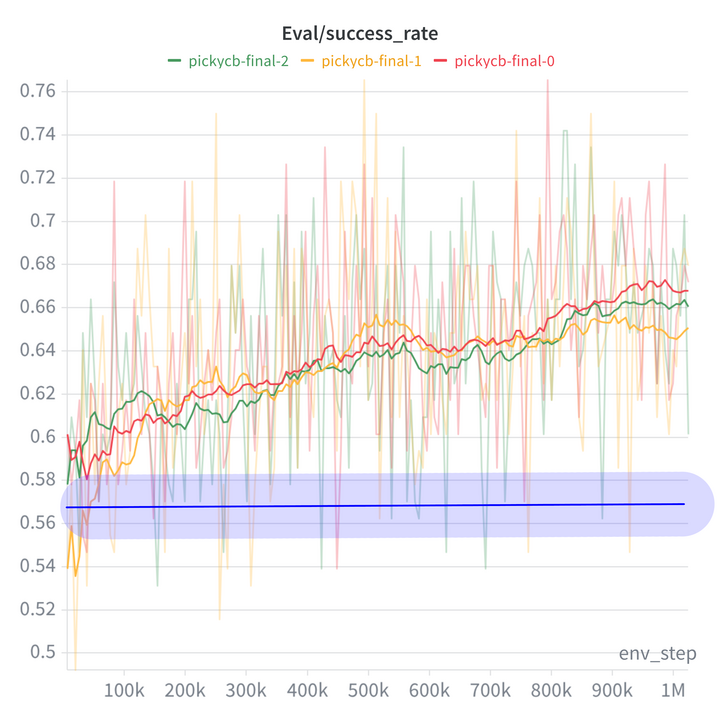}
        
        {\small \textbf{PickSingleYCB: End}}
      \end{minipage}
      \hfill
      \begin{minipage}[t]{0.48\linewidth}
        \centering
        \includegraphics[width=\linewidth]{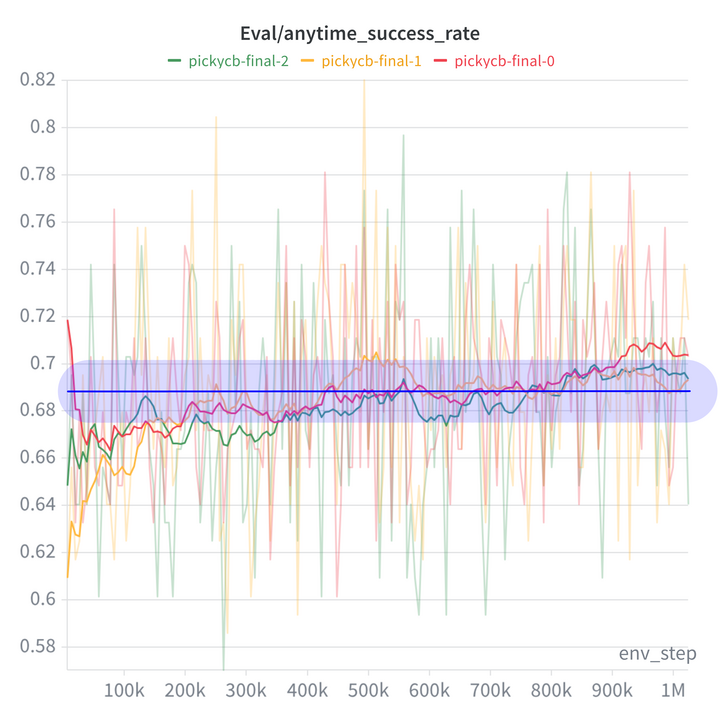}
        
        {\small \textbf{PickSingleYCB: Anytime}}
      \end{minipage}

      \caption{Success rate versus environment steps across 3 random seeds for the FM policy augmented with a learned world model. Dark curves show 0.95 EMA-smoothed trends. Blue lines indicate the FM policy baseline with 95\% confidence intervals.}
      \label{fig:success_curves}
    \end{figure}

Figure~\ref{fig:success_curves} and Tables~\ref{tab:end_success} and \ref{tab:anytime_success} compare the FM policy against the same policy augmented with a learned world model and MPPI planning. Across both ManiSkill tasks, adding the world model improves performance over the FM baseline on end-of-episode success and improves anytime success on the \texttt{PickCube} task.

The strongest gains appear in end success. On \texttt{PickCube}, end success increases from 93.14\% \(\pm\) 0.77 to 97.44\% \(\pm\) 0.48, and on \texttt{PickSingleYCB}, it increases from 56.81\% \(\pm\) 1.52 to 66.41\% \(\pm\) 1.45. These improvements suggest that planning with the learned world model is particularly helpful for correcting small action errors and maintaining control through the end of the episode, rather than only reaching a temporary success state.

Anytime success also improves on both tasks, though the effect is smaller. On \texttt{PickCube}, anytime success rises from 95.78\% \(\pm\) 0.62 to 98.68\% \(\pm\) 0.35, while on \texttt{PickSingleYCB} it increases from 68.77\% \(\pm\) 1.42 to 69.78\% \(\pm\) 1.41. Notably, the \texttt{PickSingleYCB} anytime metric is the only case where the 95\% confidence intervals overlap, indicating that the clearest benefit of the world model on that task is in sustained end-of-episode success rather than transient task completion.

On \texttt{PickSingleYCB}, the FM baseline already performs close to the approximately 60\% result reported for TD-MPC2, while the addition of the learned world model exceeds that reported level of performance. Overall, these results show that world-model-based planning provides a consistent benefit on top of a fixed FM policy. They also suggest that the benefit is largest in settings where precise final control matters.
        
    \begin{table}[h]
    \centering

    \begin{minipage}[t]{0.48\linewidth}
    \centering
    \small
    \begin{tabular}{lcc}
    \toprule
    & PickCube & PickSingleYCB \\
    \midrule
    FM policy & 93.14 $\pm$ 0.77 & 56.81 $\pm$ 1.52 \\
    \textbf{FlowMPC} & \textbf{97.44 $\pm$ 0.48} & \textbf{66.41 $\pm$ 1.45} \\
    \bottomrule
    \end{tabular}

    \vspace{0.2em}
    \captionof{table}{End success rate (\%, mean $\pm$ 95\% CI) evaluated over 4096 episodes.}
    \label{tab:end_success}
    \end{minipage}
    \hfill
    \begin{minipage}[t]{0.48\linewidth}
    \centering
    \small
    \begin{tabular}{lcc}
    \toprule
    & PickCube & PickSingleYCB \\
    \midrule
    FM policy & 95.78 $\pm$ 0.62 & 68.77 $\pm$ 1.42 \\
    \textbf{FlowMPC} & \textbf{98.68 $\pm$ 0.35} & \textbf{69.78 $\pm$ 1.41} \\
    \bottomrule
    \end{tabular}

    \vspace{0.2em}
    \captionof{table}{Anytime success rate (\%, mean $\pm$ 95\% CI) evaluated over 4096 episodes.}
    \label{tab:anytime_success}
    \end{minipage}

    \end{table}

\section{Discussion}

I believe the clearest takeaway from these results is that the world model helped most in the part of the task that is easiest to get almost right and still fail: the final execution. On both \texttt{PickCube} and \texttt{PickSingleYCB}, the gains were larger for end-of-episode success than for anytime success, which suggests that planning was not just helping the policy reach successful states, but helping it stay there through the end of the episode. In these manipulation tasks, that matters a great deal, since success requires not only moving the object to the goal but maintaining stable control afterward \citep{taomaniskill3}.

I think this is also why the overall approach feels promising for flow-based imitation policies in particular. The FM policy already provides a strong multimodal action prior, allowing for the use of binary/semi-sparse rewards, but by itself it is still limited by the demonstrations it was trained to imitate \citep{pmlr-v305-jiang25a}. Adding the world model gave the policy a way to evaluate and refine candidate trajectories at test time without changing the FM objective itself, which seems to be enough to improve performance beyond imitation alone on both tasks. To me, that is the most interesting part of the result: the improvement did not come from retraining the policy with a new reinforcement learning objective, but from making better use of the policy at inference time.

At the same time, I do not think these results are broad enough to support especially strong claims. The work only evaluates two ManiSkill tasks, and the best-performing world model design was different in each case: \texttt{PickCube} favored the simpler state-space model with \(H=1\), while \texttt{PickSingleYCB} benefited from the latent model with \(H=3\). That makes me think the success of the method depends in a meaningful way on representation choice and planning horizon, rather than on a single universally good configuration. The approach looks viable, but it is still not clear how robust these design choices would be across a wider range of tasks.

One other interesting point is that the FM baseline by itself was already fairly strong, especially on \texttt{PickCube}, and on \texttt{PickSingleYCB} it was already close to the roughly 60\% result reported for TD-MPC2 \citep{hansen2024tdmpc2scalablerobustworld}. The fact that the world-model-augmented version still improved on top of that baseline makes the result more convincing to me, because the planner was not rescuing a weak policy so much as sharpening a strong one. Future work should test whether the same pattern holds on a broader set of environments and whether other world model architectures or planning algorithms would lead to larger gains.

\section{Conclusion}

This paper's FlowMPC framework successfully leveraged a learned world model to improve the performance of a Flow Matching policy through MPPI-based planning. Across both evaluated tasks, this led to higher success rates than the FM policy alone, with especially clear gains in end-of-episode success.

I believe these results point to a promising direction for combining imitation learning and model-based reinforcement learning. FM provides a strong generative prior over action trajectories, and the world model provides a mechanism for refining those trajectories at inference time. While this work only evaluates two tasks, it offers evidence that flow-based policies can benefit meaningfully from world-model-based planning without requiring a new reinforcement learning objective.

\newpage

\bibliographystyle{plainnat}
\bibliography{reference}

\end{document}